\documentclass[conference,10pt,twocolumn]{IEEEtran}
\IEEEoverridecommandlockouts
\newcommand{\subparagraph}{}

\newif\ifdraft
\draftfalse

\ifdraft
\usepackage[letterpaper,paperwidth=9in,paperheight=11in,
			top=0.75in,bottom=1in,left=0.875in,right=0.875in,
			footskip=.25in,heightrounded,marginparwidth=0.9in,
			marginparsep=0.01in]{geometry}
\usepackage{xcolor}
\usepackage{xargs} 
\usepackage[textsize=footnotesize]{todonotes}
\newcommandx{\td}[2][1=]{\todo[inline,linecolor=blue,
			backgroundcolor=blue!10,bordercolor=blue,#1]{#2}}
\newcommandx{\jj}[2][1=]{\todo[linecolor=red,
			backgroundcolor=red!10,bordercolor=red,#1]{{\bf JJ}: #2}}
\def\ct#1{\textcolor{red}{#1}}
\else
\def\ct#1{}
\def\td#1{}
\def\jj#1{}
\fi

\usepackage{tabularx}
\usepackage{amssymb,amsmath}
\usepackage{epsfig}
\usepackage{textcomp}
\usepackage{epstopdf}
\usepackage{url}
\usepackage{wrapfig}
\usepackage{enumerate}
\usepackage{soul}
\usepackage{tipa}
\usepackage[percent]{overpic}

\usepackage[english]{babel}
\usepackage{blindtext}

\usepackage[linesnumbered,ruled]{algorithm2e}
\SetKwInput{KwIn}{Initial condition}
\SetKwInput{KwResult}{Outcome}
\SetAlgorithmName{Strategy}{List of strategies}

\usepackage{amsthm}

\usepackage{mathtools,xparse}

\usepackage{xspace}
\def\mvp{{\sc microMVP}\xspace}

\title{
\vspace*{10mm}
A Portable, 3D-Printing Enabled Multi-Vehicle Platform for 
Robotics Research and Education
}
\author{Jingjin Yu \qquad Shuai D. Han \qquad Wei N. Tang \qquad Daniela Rus%
\thanks{J. Yu, S.D. Han, and W.N. Tang are with the Department of Computer 
Science, Rutgers University at New Brunswick. E-mails: 
\{jingjin.yu, sh1067, wt160\}@cs.rutgers.edu. D. Rus is with the Computer Science 
and Artificial Intelligence Lab, the Massachusetts Institute of Technology. 
E-mail: rus@csail.mit.edu.}
\thanks{We thank Aaron Becker and the reviewers for their helpful comments. 
This work was supported in part by NSF grant 1617744 (IIS Robust Intelligence), 
a Rutgers Research Council grant, and ONR projects N00014-12-1-1000 and
 N00014-09-1-1051.}%
}

\begin{document}
\maketitle
\jj{The .tex file can be compiled with standard latex distributions. For 
windows, you may use miktex (2.9+) and texniccenter. To use side comments, 
use $\backslash$td\{text\}. Pages are wider than they should in draft mode.
To get the final compilation for submission or to do normal printing, 
replace {\em drafttrue} with {\em draftfalse} and recompile.}
\begin{abstract}
\mvp is an affordable, portable, and open source micro-scale mobile 
robot platform designed for robotics research and education. 
As a complete and unique multi-vehicle platform enabled by 3D printing and 
the maker culture, \mvp can be easily reproduced and requires little 
maintenance: a set of six micro vehicles, each measuring $8\times 5\times 
6$ cubic centimeters and weighing under $100$ grams, and the accompanying 
tracking platform can be fully assembled in under two hours, all from 
readily available components. In this paper, we describe \mvp's hardware 
and software architecture, and the design thoughts that go into the making 
of the platform. The capabilities of \mvp APIs are then demonstrated with 
several single- and multi-robot path and motion planning algorithms. 
\mvp supports all common operation systems. 
\end{abstract}

\section{Introduction}
\begin{figure}[ht!]
\begin{center}
\begin{overpic}[width=0.48\textwidth,tics=20]
{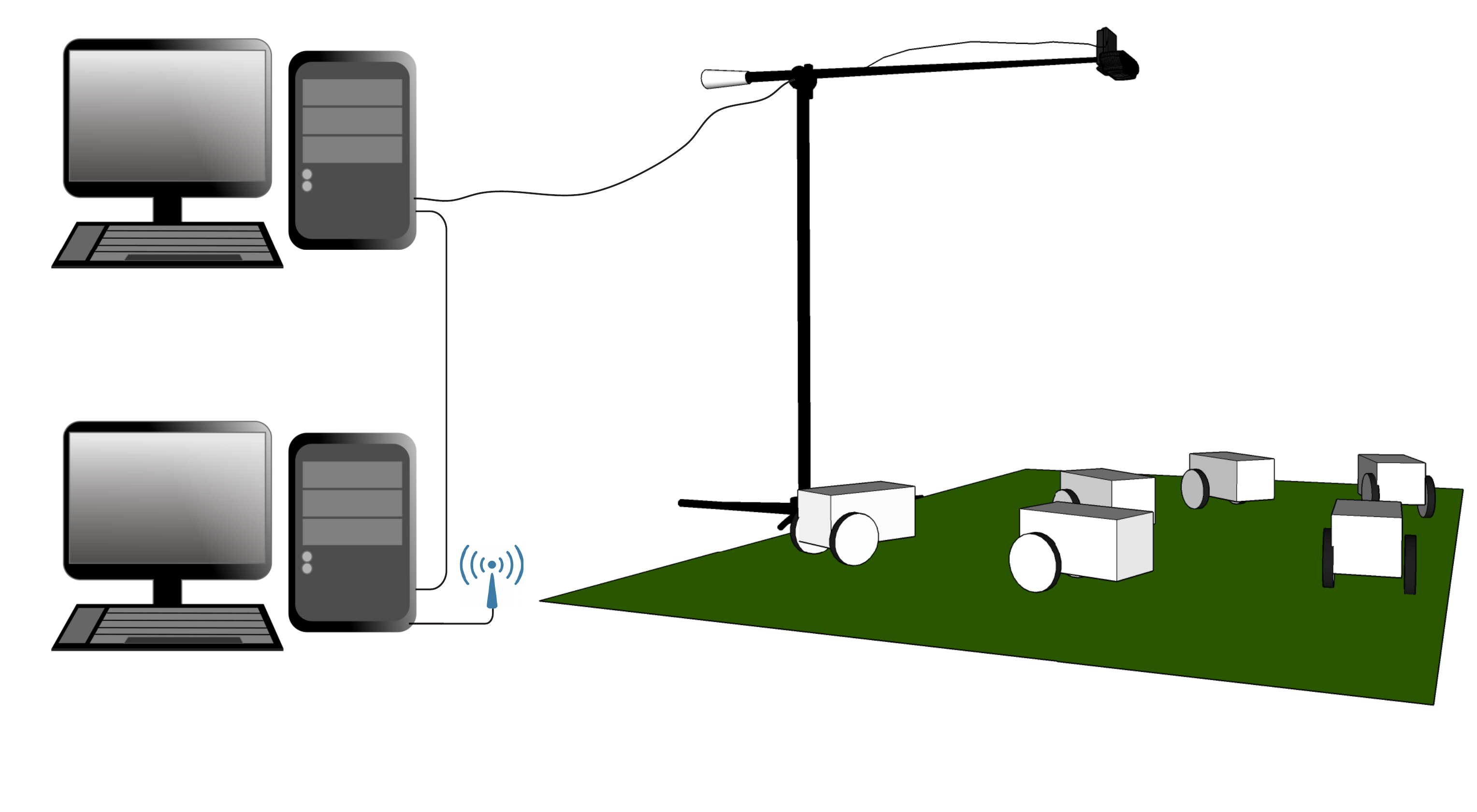}
\put(8,31){{\small Vehicle ID}}
\put(5,5){{\small Vehicle control}}
\put(56,45){{\small State tracking platform}}
\put(60,26){{\small Micro vehicle swarm}}
\end{overpic} \\
(a) \vspace*{2mm}\\
\includegraphics[width=0.48\textwidth]{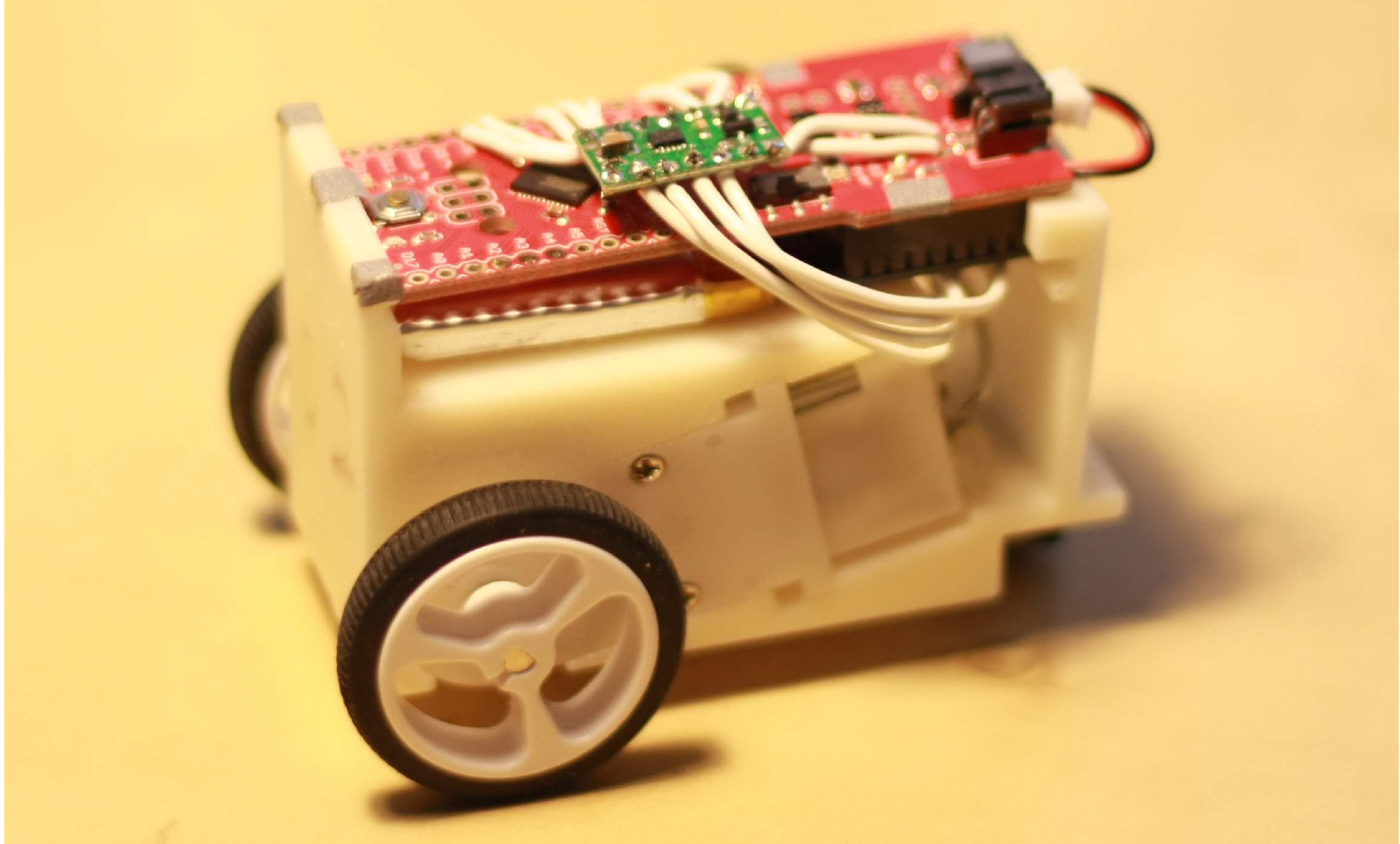}\\
(b)
\end{center}
\vspace*{-2mm}
\caption{(a) An illustrative model of \mvp platform architecture. 
Control signals are delivered to the vehicles wirelessly. The vehicle 
control portion of the system is where users can inject desired 
control logic. (b) A single fully assembled vehicle, measures $8 cm (l)\times
5 cm (w) \times 6 cm (h)$.}
\label{fig:mvp}
\vspace*{-8mm}
\end{figure}

In this paper, we introduce an affordable, portable, and open source 
multi-vehicle hardware and software platform, \mvp\footnote{We call 
the overall system as \mvp, standing for {\em 
micro-scale Multi-Vehicle Platform}.} (Fig.~\ref{fig:mvp} (a)), for research 
and education efforts requiring single or multiple mobile robots. \mvp 
consists of highly compact micro-vehicles, a state tracking camera system, 
and a supporting software stack. Each micro-vehicle (Fig.~\ref{fig:mvp} (b)) 
has a rigid 3D-printed shell allowing the precise (snap-on) fitting of the 
essential components--the Arduino-based sensorless vehicle measures less than 
$8$-cm in length and can be controlled wirelessly at a frequency of over $100$Hz. 
The external state-tracking system consists of a single USB webcam for 
estimating the configurations of the vehicles (in $SE(2)$). The overall
system is capable of feedback control of the entire vehicle fleet at a 
control update frequency of $30$Hz and above. The components in each vehicle
costs less than 90 USD and the tracking platform costs about 80 USD. We 
expect the cost of \mvp to drop significantly with the release of future 
iterations of the platform. 

We develop \mvp with both research and education applications in 
mind. The relatively high accuracy of \mvp with respect to its state 
estimation and control capabilities renders the platform suitable as a 
testbed for single- or multi-robot path and motion planning algorithms. 
As examples, we demonstrate that \mvp can seamlessly integrate
{\em (i)} centralized multi-robot path planning algorithms 
\cite{YuLav15TRO-II} and {\em (ii)} (distributed) reciprocal velocity 
obstacle algorithms \cite{BergLinManocha08RVO,Snape2014DifferntialDrive}. 
On the educational side, the affordability and portability of \mvp makes 
it ideal for the teaching and self-education of robotics subjects 
involving mobile robots. Moreover, the platform requires little time 
commitment to reproduce and maintain. The tracking platform requires minimal 
setup and each vehicle can be built in under 20 minutes. 

{\bf Related work and differentiation}. Since a large number of mobile 
robots have been produced with various capabilities and we cannot hope to 
enumerate them all, here, we focus on recent research- and 
education-centered ground mobile robot platforms intended for multi-robot 
coordination and collaboration tasks. Due in part to the rapid advances 
in MEMS technology in recent years and the related maker 
movement\footnote{\url{https://en.wikipedia.org/wiki/Maker_culture}}, it 
becomes increasingly feasible for robotics researchers, educators, and 
hobbyists alike to produce highly capable mobile robots at lower costs. 
Representative ones include the e-pucks educational robots 
\cite{mondada2009puck}, kilobots collective mobile robots
\cite{rubenstein2014kilobot}, the duckiebot autonomous mobile 
robots\footnote{\url{http://duckietown.mit.edu/}}, and 
Robotarium \cite{pickem2016safe}. 
The e-puck robots \cite{mondada2009puck} are differentially driven robots (DDR)
with two independent motor thrust input. An e-puck robot measures $7cm$ in 
diameter and weighs about $200$ grams. It hosts an array of sensors including 
microphone arrays, proximity sensors, accelerometers, and so on. 
Kilobots \cite{rubenstein2014kilobot} are smaller coin sized robots that 
locomote via vibration (over smooth surfaces), making them ideal for 
experimenting with collective behavior that is frequently found in nature
\cite{Rey87}. The duckiebots present a recent attempt from MIT for teaching 
students about autonomous driving with hands on experience. Robotarium is 
a recent NSF funded effort at GeorgiaTech that is intended as a 
remotely accessible, extensive robotics testbed, which includes a multi-vehicle
platform with inch-sized DDR vehicles.  

\begin{figure}[htp]
\begin{center}
\begin{tabular}{ccc}
\includegraphics[width=0.21\textwidth]{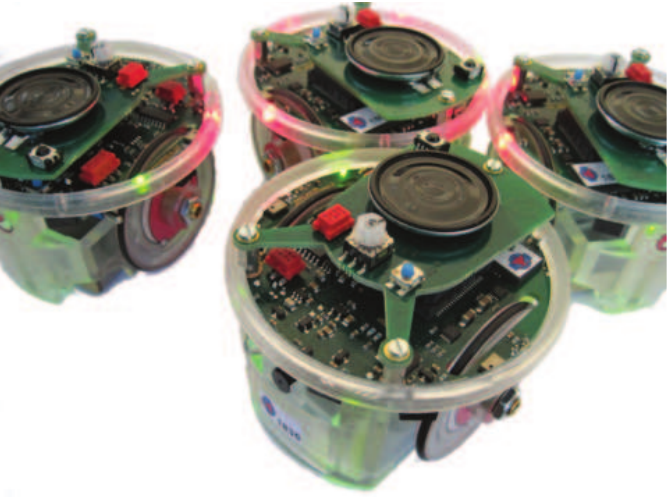} & &
\includegraphics[width=0.21\textwidth]{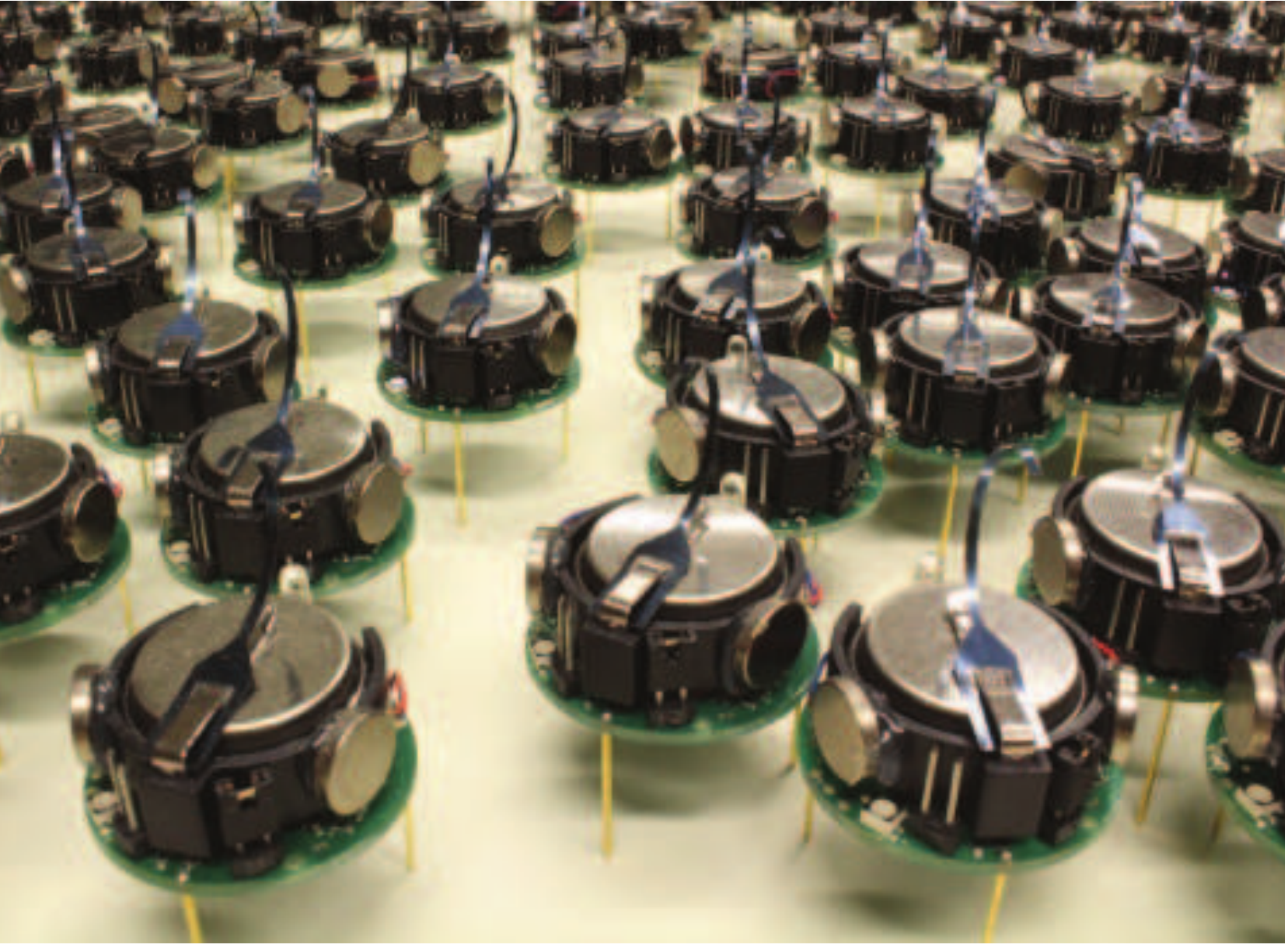} \\
(a) && (b)\\
\end{tabular}\\
\begin{tabular}{ccc}
\includegraphics[width=0.21\textwidth]{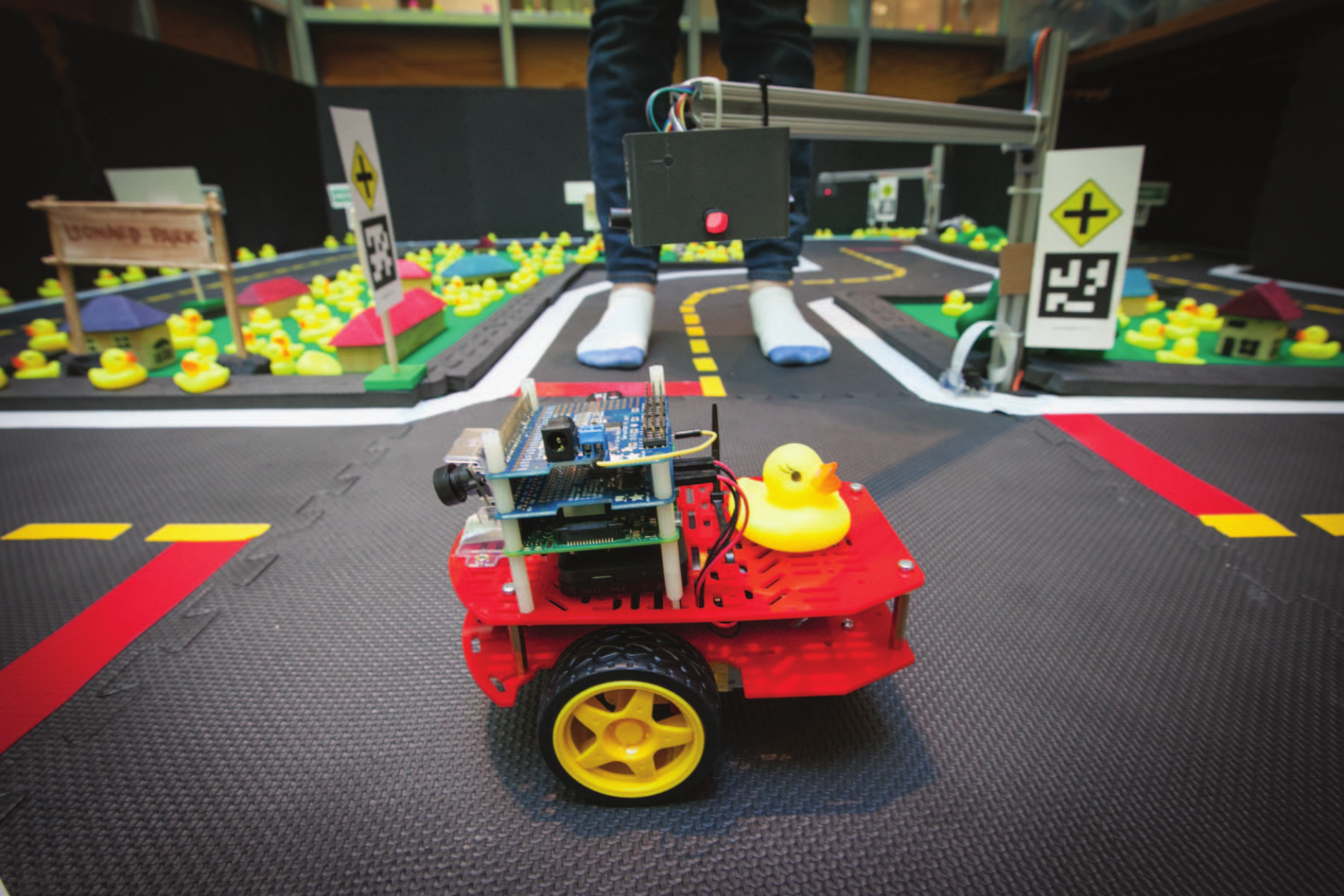} & &
\includegraphics[width=0.21\textwidth]{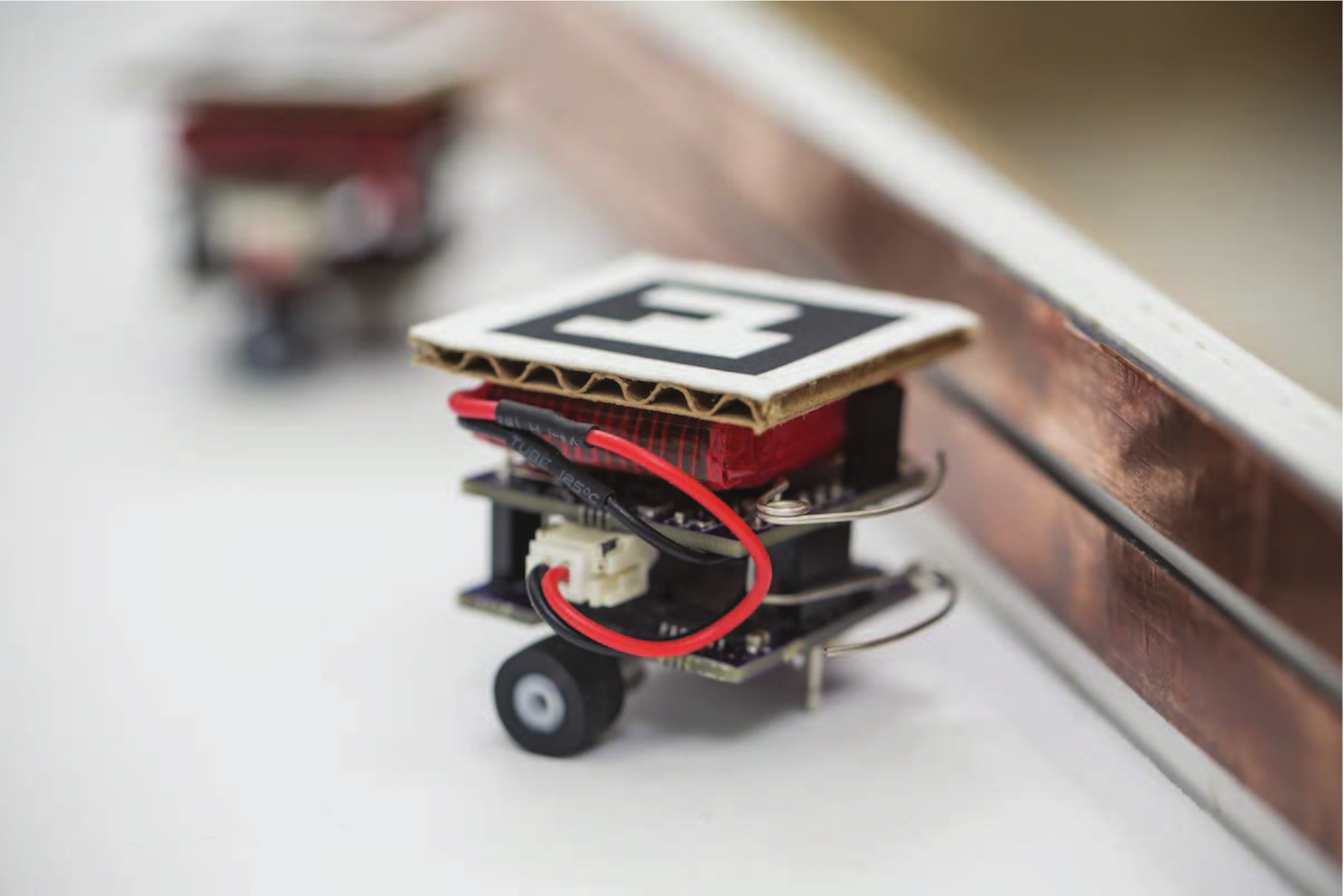} \\
(c) && (d) \\
\end{tabular}
\end{center}
\caption{A few popular mobile multi-robot platforms. (a) E-pucks ($\sim 7cm$). 
(b) Kilobots collective robots ($\sim 2cm$). (c) A duckiebot in a duckietown
($\sim 25cm$). (d) A DDR self-recharging robot in the Robotarium project ($\sim 3cm$).}
\label{fig:platforms}
\end{figure} 

Each of the above-mentioned mobile robot platforms has its particular 
strengths. However, when it comes to robot path planning, these 
platforms also have their limitations. For example, e-pucks are compact 
and capable, but they are expensive to acquire and maintain, costing 1K+ 
USD each. kilobots, great for swarm-related studies, are not ``garden variety''
mobile robots, making them unsuitable for multi-robot path planning 
experiments. Platforms like Duckiebots and Robotarium's also have their 
drawbacks (e.g., larger footprint and remote accessibility, respectively).
To address our own research and education needs, and seeing the limitations 
of existing solutions for the general task of single- and multi-robot path 
planning, we developed \mvp for filling the gap as a capable, affordable, 
portable, readily available, and low maintenance multi-vehicle platform. 

{\bf Contributions}. \mvp is a capable, simple, affordable mobile robot 
platform with a very small footprint. First and most importantly, as we
will demonstrate, \mvp is highly capable as an experimental platform for 
both centralized and distributed multi-robot planning and coordination 
tasks. Secondly, it presents a solution that is fairly portable and compact, 
suitable for showcasing multi-robot systems in action in limited space, 
making it ideal for both research and education. Last but not least, 
\mvp's open source design, utilizing 3D printing technology, is 
extremely simple and robust. Significant care is also taken to ensure
that only readily available components are used, which makes \mvp a truly 
readily reproducible mobile robot platform. 


The rest of the paper are organized as follows. In 
Section~\ref{section:hardware}, we describe the hardware architecture of \mvp,
followed by a summary of thoughts going into the design of the software 
stack in Section~\ref{section:api}. We highlight the applications and 
capabilities of \mvp in Section~\ref{section:capability} and conclude in 
Section~\ref{section:conclusion}. We note that this paper describes mainly 
\mvp's design philosophy and capabilities. Additional details of the 
system, including component acquisition, assembly instructions, API interfaces, 
examples, can be found at \url{http://arc.cs.rutgers.edu/mvp/}.

\section{Platform Architecture and Design}\label{section:hardware}
The main goal in designing \mvp is to optimally combine portability and 
availability. Availability further entails affordability, component 
availability, and low system assembly and maintenance time. To reach the 
design target, we explored a large number of micro controller families 
(Arduino, mbed, and Raspberry Pi), wireless technology (wifi, zigBee, 
and Bluetooth), motors and motor controllers, overall vehicle design (3D 
printing, laser cut machining, and off-the-shelf kits), and tracking 
technology (infrared-marker, fiducial marker). We eventually fixated on 
the design choice of using Arduino/zigBee for communication and control,
plastic gear motors for mobility, and fiducial markers for system tracking. 

\mvp has two main components: {\em (i)} a 3D-printing enabled micro-vehicle 
fleet and {\em (ii)} a fiducial marker tracking system for vehicle state 
estimation. Fig.~\ref{fig:mvp} illustrates the platform architecture 
(with $6$ vehicles, additional vehicles can be readily added without 
additional infrastructure) and provides a picture of a single fully 
assembled vehicle. The general architecture of \mvp is rather 
straightforward: the web-cam based sensing system continuously queries the 
configuration of the vehicles, upon which planning decisions are made and 
translated into control signals that are relayed wirelessly to the vehicles. 
\mvp can run the feedback control loop at a frequency of up to $100$Hz, or 
as high as limited by the frames-per-second (fps) rating of the camera's 
capture mode. For webcams, this is normally $30$fps or $60$fps. In what 
follows, we describe the design of the individual components of \mvp in 
more detail.
\subsection{3D-Printing Enabled Micro-Scale Vehicles}
\begin{figure}[htp]
\begin{center}
\begin{tabular}{ccc}
\includegraphics[width=1.8in]{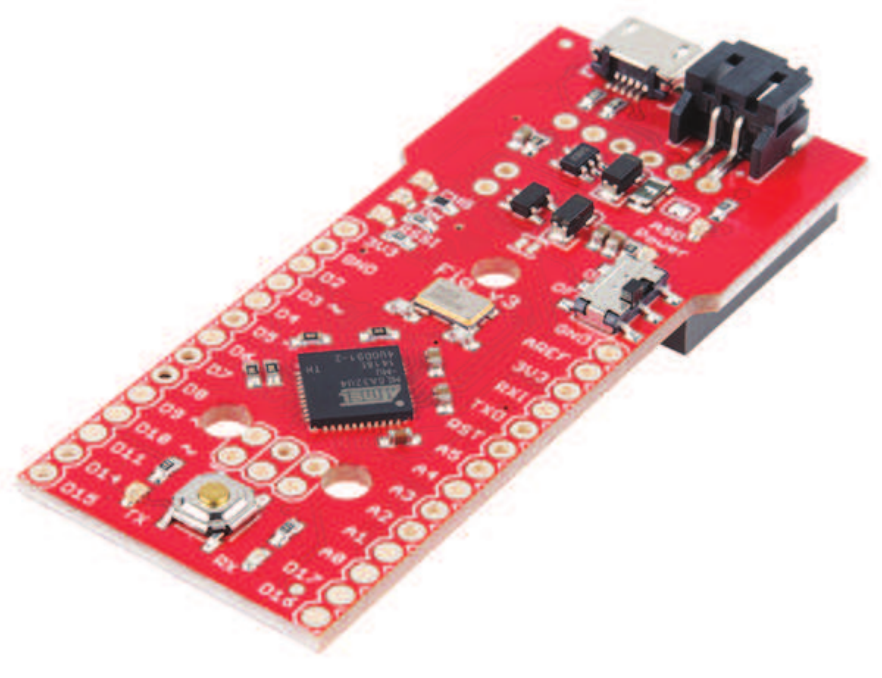} & &
\includegraphics[width=0.9in]{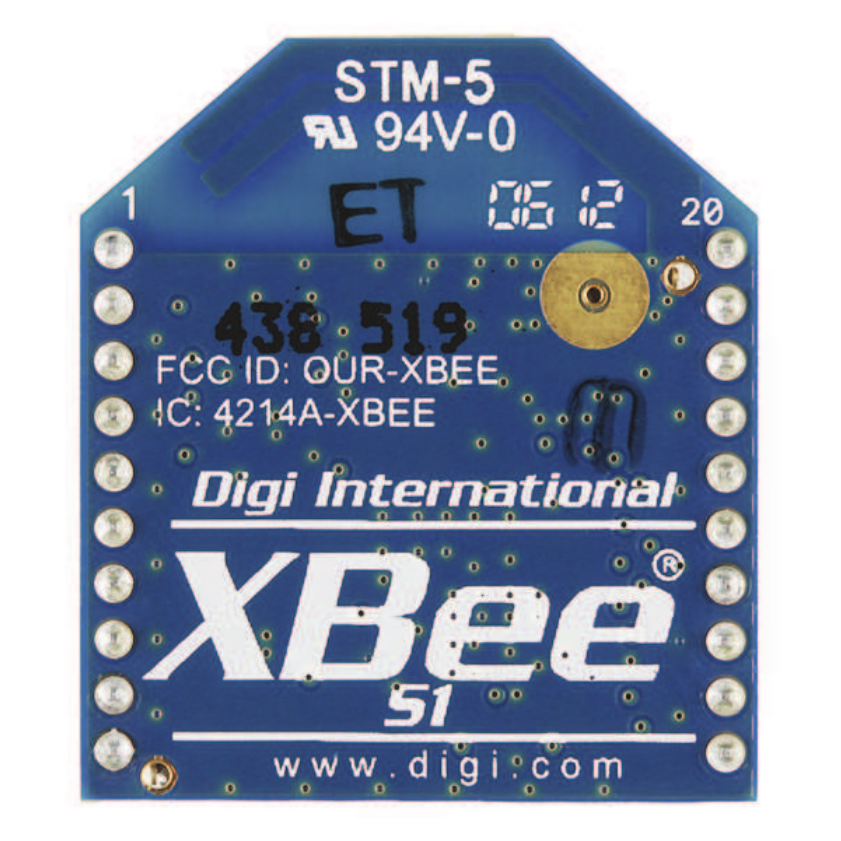} \\
(a) && (b)\\
\end{tabular}\\
\begin{tabular}{ccc}
\includegraphics[width=0.85in]{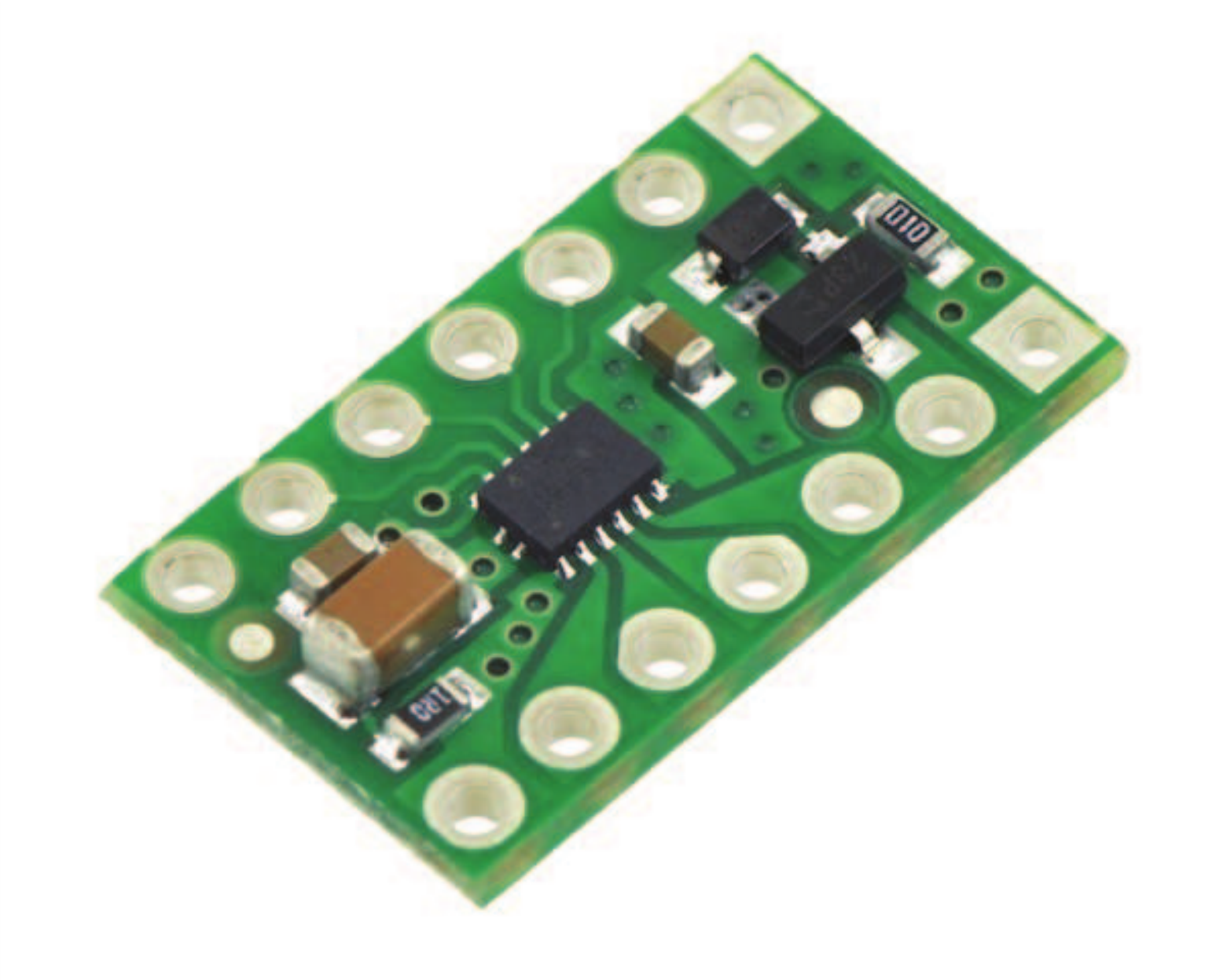} & &
\includegraphics[width=1.5in]{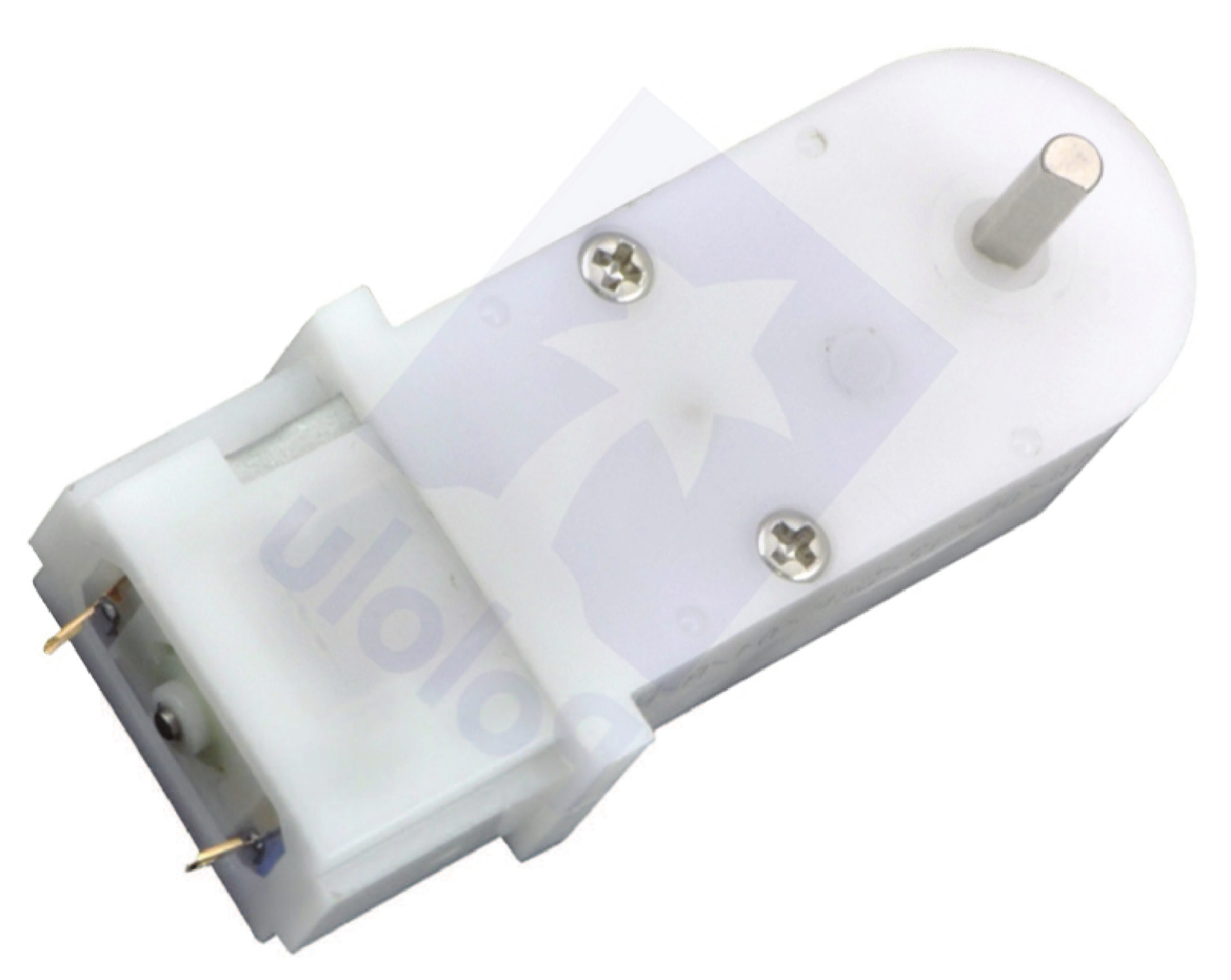} \\
(c) && (d) \\
\end{tabular}
\end{center}
\caption{Key components of the vehicle, minus the 3D printed shell, the 
battery and the wheels. (a) Sparkfun fio v3 board with built-in xBee 
support (socket on the back). (b) An XBee (Series 1) module with 
built-in trace antenna. (c) Pololu DRV8835 motor driver. (d) Pololu 
120:1 plastic gear motor.}
\vspace*{-2mm}
\label{fig:vehicle}
\end{figure}

We design the vehicle to be small, affordable, reliable, and easily
reproducible. To reach the design goal of reliability while maintaining 
simplicity, the vehicle is built around the proven zigBee Series 1 
(Fig.~\ref{fig:vehicle}(b)) as the wireless communication module (zigBees 
are frequently found in research drones). For actuation, two micro gear 
motors (Fig.~\ref{fig:vehicle}(d)) are connected to Pololu DRV8835 dual 
motor driver carrier (Fig.~\ref{fig:vehicle}(c)). Then, the zigBee module, 
the motor driver carrier, and a 400mAh 3.7v Li-ion battery are fitted to a 
Sparkfun fio v3 board (Fig.~\ref{fig:vehicle}(a)) which is Arduino 
compatible. In additional to built-in zigBee support, the fio v3 board 
allows direct charging of the battery, a convenient feature. To put things 
together, all components are snapped into a 3D printed plastic shell and 
wheels are attached to the motors. The use of 3D printing is essential in 
the design phase, allowing both the precise fitting of the components and 
a quick screw-less assembly. Finally, a small caster wheel is attached 
to reduce the friction of the vehicle, completing the vehicle hardware 
(Fig.~\ref{fig:mvp}(b)), which measures less than $8cm \times 5cm 
\times 6cm$ and weighs just below 100 grams. A fully charged battery 
allows the vehicle to operate continuously for about one and half hours. 
Recharging on a standard 0.5A USB port takes less than one hour. 

The components of the vehicle cost less than 90 USD in total (please refer 
to the project website for a list of individual components and their costs). 
Excluding the time required for 3D printing of the vehicle shell, it takes 
less than 20 minutes to assemble the vehicle. All that is required is to 
solder a number of wires, glue together a few components, and snap them 
together. We tested the vehicles over a variety of flat surfaces commonly 
found in classrooms, offices and labs, including wood floors, thin carpets, 
vinyl floors, and concrete slabs. We could verify that the vehicles do not 
need special surfaces to operate. The top speed of the vehicles is relatively
uniform on different surfaces with the exception of carpeted surface, where
the top speed is reduced as the thickness of the carpet increases. 

\subsection{Camera Platform for Vehicle Tracking}
\mvp uses a single USB camera (we selected the Logitech C920, 
$1920\times1080$ max resolution) for tracking the $SE(2)$ configurations 
of the vehicles. The camera is mounted to a microphone stand with a tripod 
base and adjustable height. The cost-effective setup, with the camera 
mounted at a height about $1m$ and facing down, establishes an approximately 
$1.5m\times 0.9m$ rectangular workspace for the vehicles at the ground level. 
The camera platform can comfortably accommodate $1$ to $20$ vehicles 
depending on the application. For vehicle tracking, after evaluating 
several open source fiducial marker based tracking package, we opted for 
chilitags\footnote{\url{https://github.com/chili-epfl/chilitags}} as the 
baseline platform for tracking vehicle states. This requires the fixture of 
markers on the vehicles, for which we used a marker size of $3.5cm\times 3.5cm$ 
(without the extra white borders; some borders must be included for good 
tracking quality). 

\subsection{Computation Hardware}
As illustrated in Fig.~\ref{fig:mvp}, \mvp system requires two pieces 
of enabling computation: tracking and control. We note that 
Fig.~\ref{fig:mvp} includes two computers to suggest that the system is 
rather modular, i.e., the tracking and the control are easily separated. 
The computation, however, can be carried out using a single multi-core 
commodity PC. 

\section{Software Stack and API}\label{section:api}
Our design on the software side of \mvp seeks to maximize modularity, 
simplicity, and cross-platform compatibility. Modularity enables 
flexibility in hardware setup and more importantly, extends the capability 
of the system. Simplicity and cross-platform compatibility are essential 
for any system aimed at mass adoption, especially for educational purposes. 
Below, we briefly describe how \mvp achieves these design goals in its 
software stack and application programming interface (API) implementation. 
For cross-platform and rapid development support, we chose Python as the 
language for developing the API, which has a state estimation component and
a vehicle control component. 

\subsection{Vehicle State Estimation}
\begin{figure}[ht!]
\begin{center}
\includegraphics[width=0.48\textwidth]{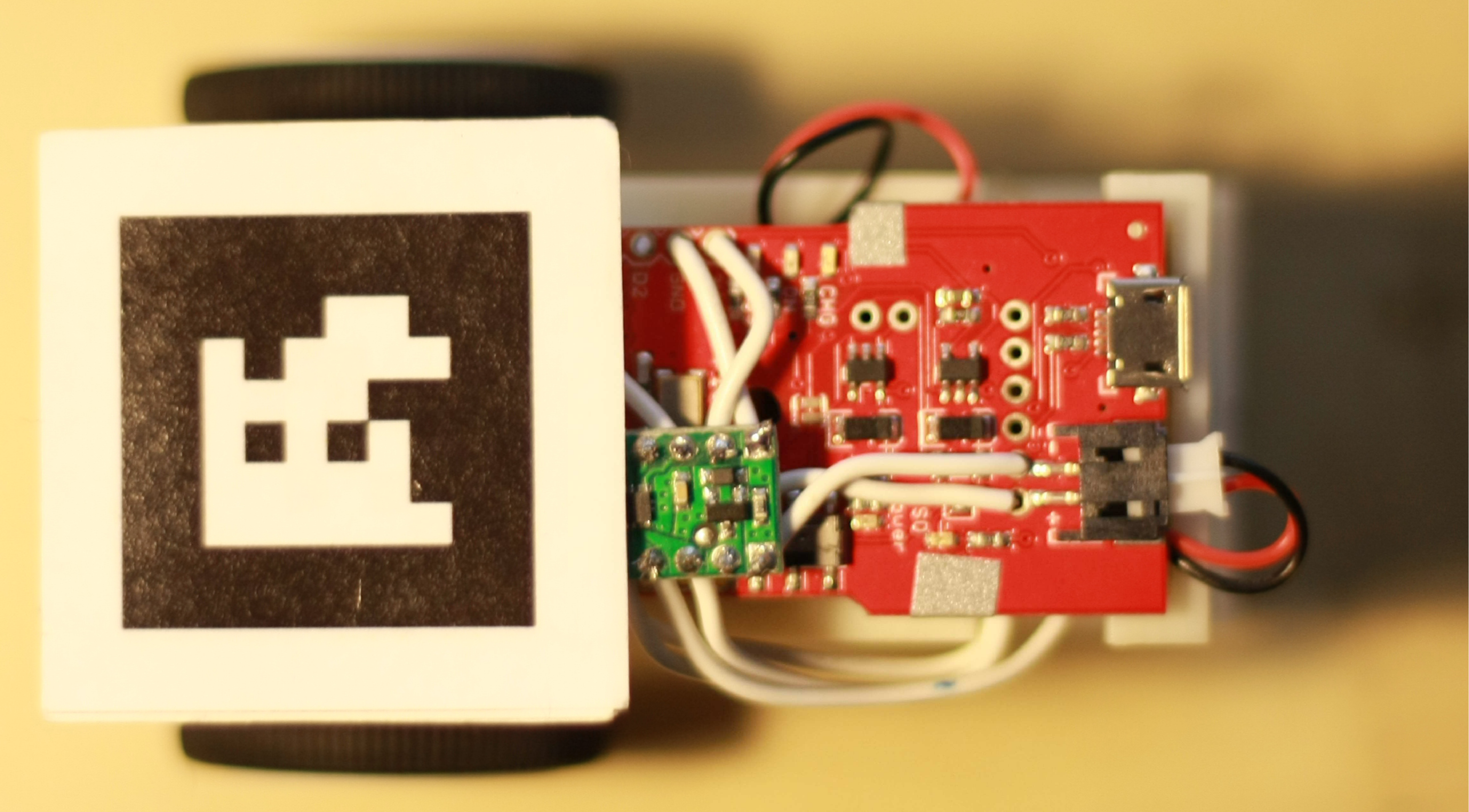}
\end{center}
\caption{A vehicle with a chilitags fiducial marker (tag) attached.}
\label{fig:vehicle-tag}
\end{figure}

For state estimation, to extract the configurations of the vehicles in 
$SE(2)$ (i.e., $\mathbb R^2 \times S^1$), we attach a $3.5cm\times 3.5cm$ 
chilitags fiducial marker on the top of the micro vehicles so that the front
side of the tag is aligned with the front of the vehicle and the center of tag 
overlaps the mid point between the two motor axles (see 
Fig.~\ref{fig:vehicle-tag}). 

The chilitags library provides open source APIs for extracting tags locations 
from images. The reported tag location (e.g., the $(x, y)$ coordinates of 
the four corners of a tag within a raster image), combined with known 
physical size of the tag and known camera orientation/calibration, allows the
estimation of the $SE(3)$ position and orientation of the tags. As vehicles 
in \mvp live on the floor, there is no need to extract the $SE(3)$ 
configurations of the tags; $SE(2)$ is sufficient. For continuous image 
acquisition through the USB camera, one may use OpenCV (Open Source Computer 
Vision Library) \cite{itseez2015opencv}. With a calibrated camera and the 
right parameters (e.g., exposure, white balance, and so on), we are able to 
track 50 tags simultaneously at 30 frames per second and rarely miss any of 
the tags. It appears that the limiting factor with respect to the frame rate
of the system mainly hinges on the frame rate of the USB camera. 

To allow user of \mvp access to the tags' 2D configurations, we adapt a 
queuing architecture much like ``topics'' in the Robotic Operation System 
(ROS). More specifically, we use a publisher-subscriber model from ZeroMQ 
\footnote{\url{http://zeromq.org/}} to deliver vehicle position data over 
the network. An end user may asynchronously request in Python, in a 
thread-safe manner, the latest vehicle positions in $SE(2)$ either for all 
vehicles or for a subset of vehicles.

\subsection{Vehicle Control}
Suppose that an end user has made control decisions for each vehicle in 
a vehicle swarm, in the form of thrusts for the wheels. \mvp provides a 
Python interface that allows the direct delivery of such control input
to the desired vehicle. To realize this, a simple data transfer protocol 
is developed to operate over an automated zigBee link between the control 
computer and the individual vehicles. The zigBee modules all operate in 
the same mode and form a communication graph that is a complete graph, 
although the effective communication graph is a directed star graph because 
vehicles only receive data from the control computer and do not send any 
data. On the vehicle side, the Arduino-based Fio v3 board continuously 
monitor the zigBee serial interface for new input. We found that the 
zigBee interface can sometimes become unreliable, a behavior that must be
compensated through software. Through a careful implementation, we were 
able to reliably deliver motor thrusts to 14 vehicles at a frequency of 
50Hz. A feedback loop running at 10Hz is generally sufficient 
for controlling these micro vehicles. 

In addition to basic control APIs that allow sending 
motor thrusts to the vehicles, \mvp also supplies many high level APIs 
that relieve the user from tedious tasks such as synchronous path following
(see Section~\ref{section:capability}). 

\subsection{Additional Features and Extensions}
\mvp comes with many additional features and extensions. We mention a few 
such possibilities here. 

{\em Multi-platform support.} In the development of \mvp, we 
made a conscious effort to build a software stack that is inherently 
multi-platform friendly. In addition to selecting Python as the language 
for development, the libraries that we use all have cross-platform
support. 

{\em Vehicle simulation.} The vehicle that we developed is
inherently a icra-2017
ly driven robot, the configuration transition 
equation of which may be represented as \cite{Lav06}
\begin{equation}\label{ddr}
\begin{array}{l}
\displaystyle\dot x = \frac{\strut r}{2}(u_l + u_r)\cos \theta\\
\displaystyle\dot y = \frac{\strut r}{2}(u_l + u_r)\sin \theta\\
\displaystyle\dot \theta = \frac{\strut r}{L}(u_r - u_l),
\end{array}
\end{equation}
in which $r$ is the wheel radius, $u_r$ and $u_l$ are the control input to 
the right and left wheels, respectively, and $L$ is the distance between 
the two wheels' centers. $(x, y, \theta)$ is the $SE(2)$ configuration of 
the vehicle. 

It is relatively straightforward to simulate other vehicle motion models 
with DDR. As an example, looking at Dubin's Car \cite{Lav06}, e.g., 
\begin{equation}\label{dubins}
\begin{array}{l}
\displaystyle\dot x = v_0\cos \theta\\
\displaystyle\dot y = v_0\sin \theta\\
\displaystyle\dot \theta = \omega_0u, \qquad u \in [-1, 0, 1],
\end{array}
\end{equation}
in which $v_0$ and $\omega_0$ are some constant line and angular velocity,
respectively, we observe that~\eqref{ddr} may readily simulate~\eqref{dubins}.
Similarly, other vehicle types including Reeds-Shepp car, simple car, 
unicycle, can all be simulated using a DDR vehicle. 

{\em Enhanced portability.} The entire \mvp platform can be put 
in a small box minus the camera stand. It is
possible to run both vehicle tracking and vehicle control on the same 
computer, suggesting that the entire system is highly portable. 
Alternatively, one may choose to offload the vehicle tracking computation
to a dedicated system with minimal footprint. As a proof of concept, we 
have tested running the system on a Raspberry Pi 2 model B and confirm
it has full functionality. 

{\em ROS integration.} \mvp, given that its source is open, can be readily 
adapted to work with ROS. In particular, the ZeroMQ based message queuing 
can be replaced with ROS topics. Similarly, vehicle control can also be 
packaged into a topic-based launch module. Due to the relatively large 
footprint of ROS installation and our design goals, \mvp is made independent 
of ROS. 

\section{Capabilities and Applications of \mvp}\label{section:capability}
\begin{figure}[ht!]
\begin{center}
\includegraphics[width=0.47\textwidth]{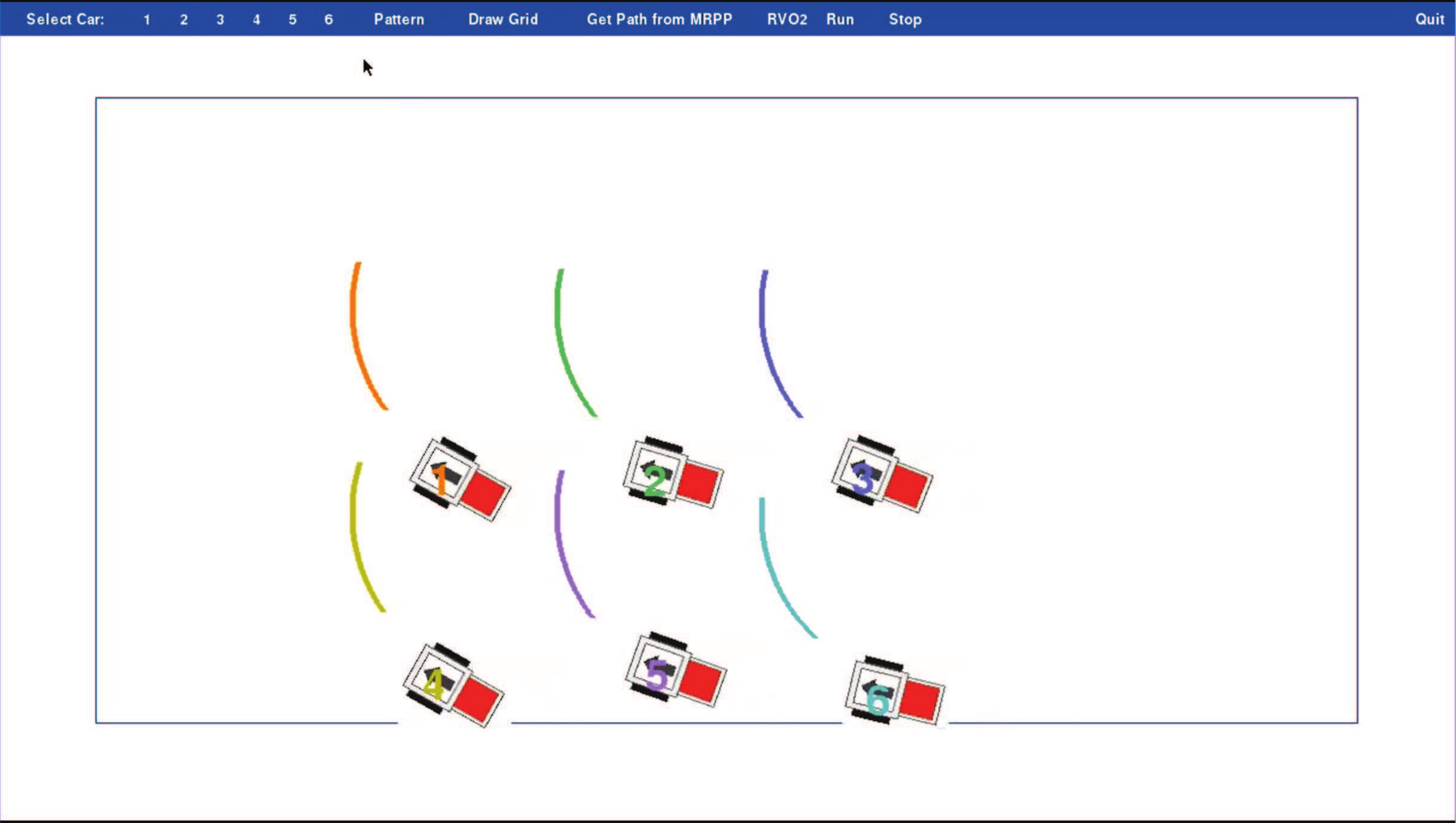}\vspace*{2mm}\\
(a) \vspace*{2mm}\\
\includegraphics[width=0.47\textwidth]{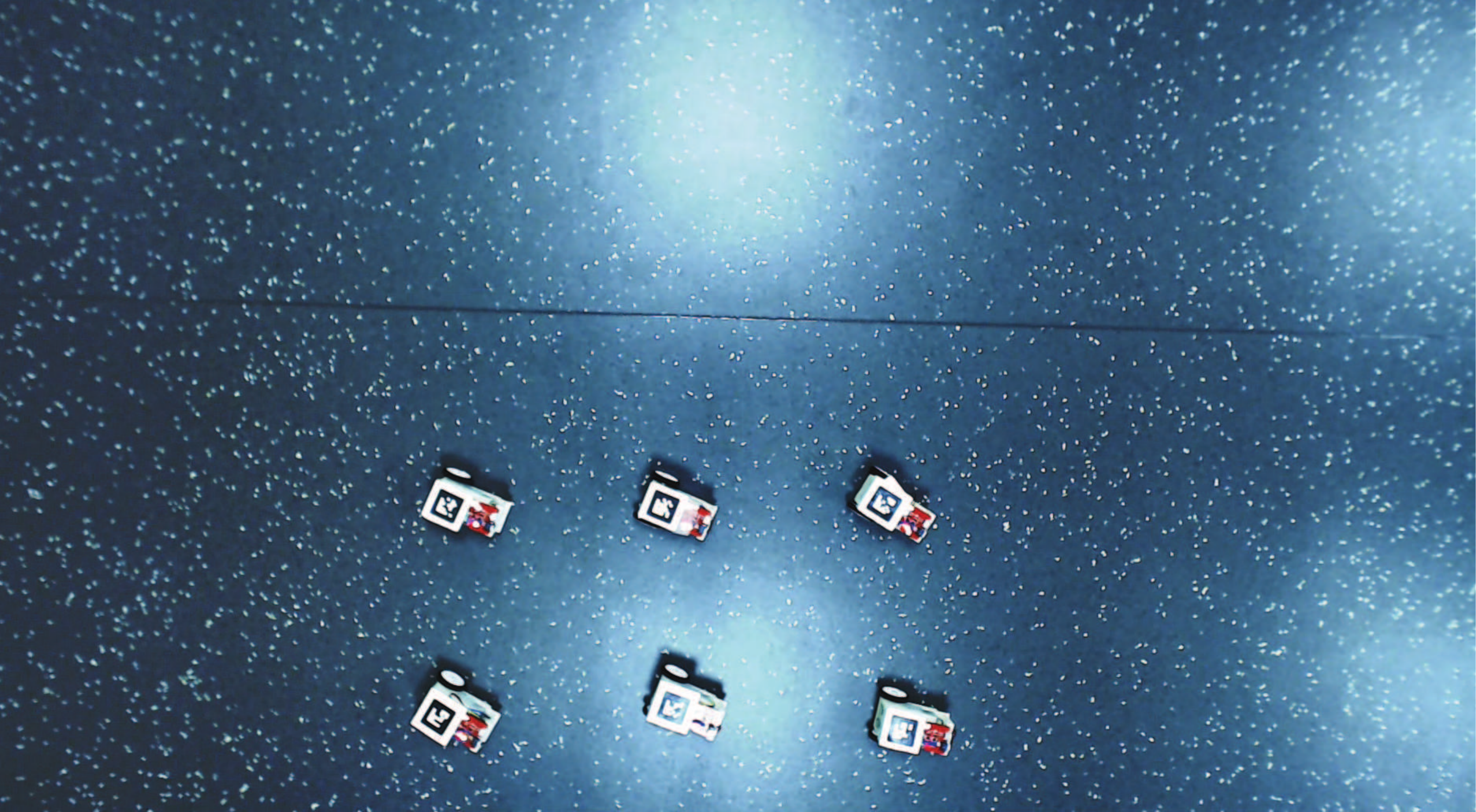}\\
(b)
\end{center}
\caption{A snapshot of the Python GUI (a) and \mvp hardware platform 
(b) in action with six moving vehicles. Note that significant glare
is present, which \mvp can deal with quite well.}
\label{fig:gui}
\end{figure}

After describing the hardware and software structures of \mvp, we provide 
several application scenarios demonstrating the capability of \mvp. We note 
that some of these capabilities, for example path following, are also part 
of \mvp APIs that an end user may readily use. A Python-based control 
graphical user interface (GUI) (see Fig.~\ref{fig:gui} for a snapshot of 
the GUI and the the corresponding hardware experiment in action)
tracks and displays the locations of the vehicles in real-time. 

\begin{figure}[ht!]
\begin{center}
\begin{tabular}{ccc}
\raisebox{-.5\height}
{\includegraphics[width=0.40\textwidth]{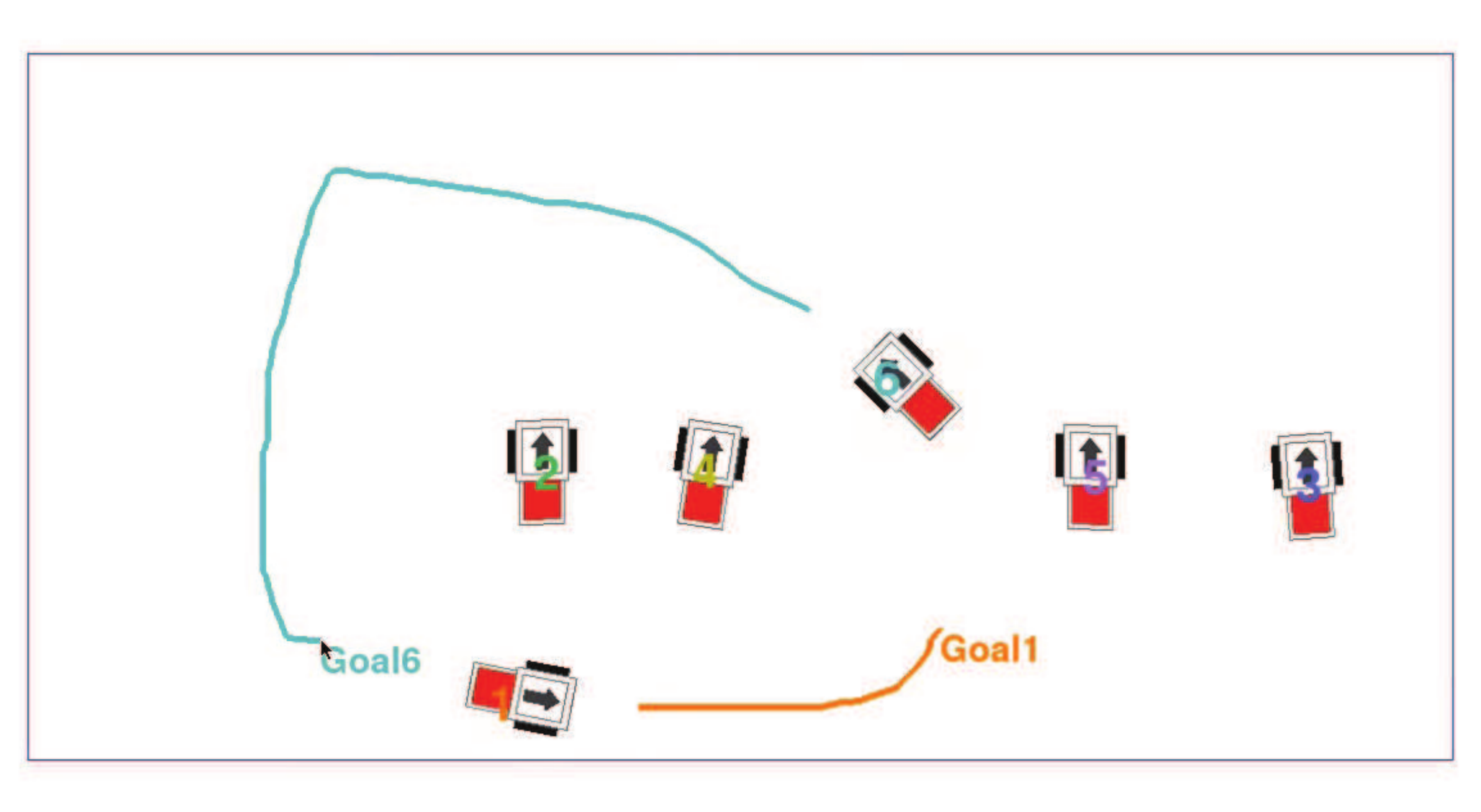}} && (a) \\
\raisebox{-.5\height}
{\includegraphics[width=0.40\textwidth]{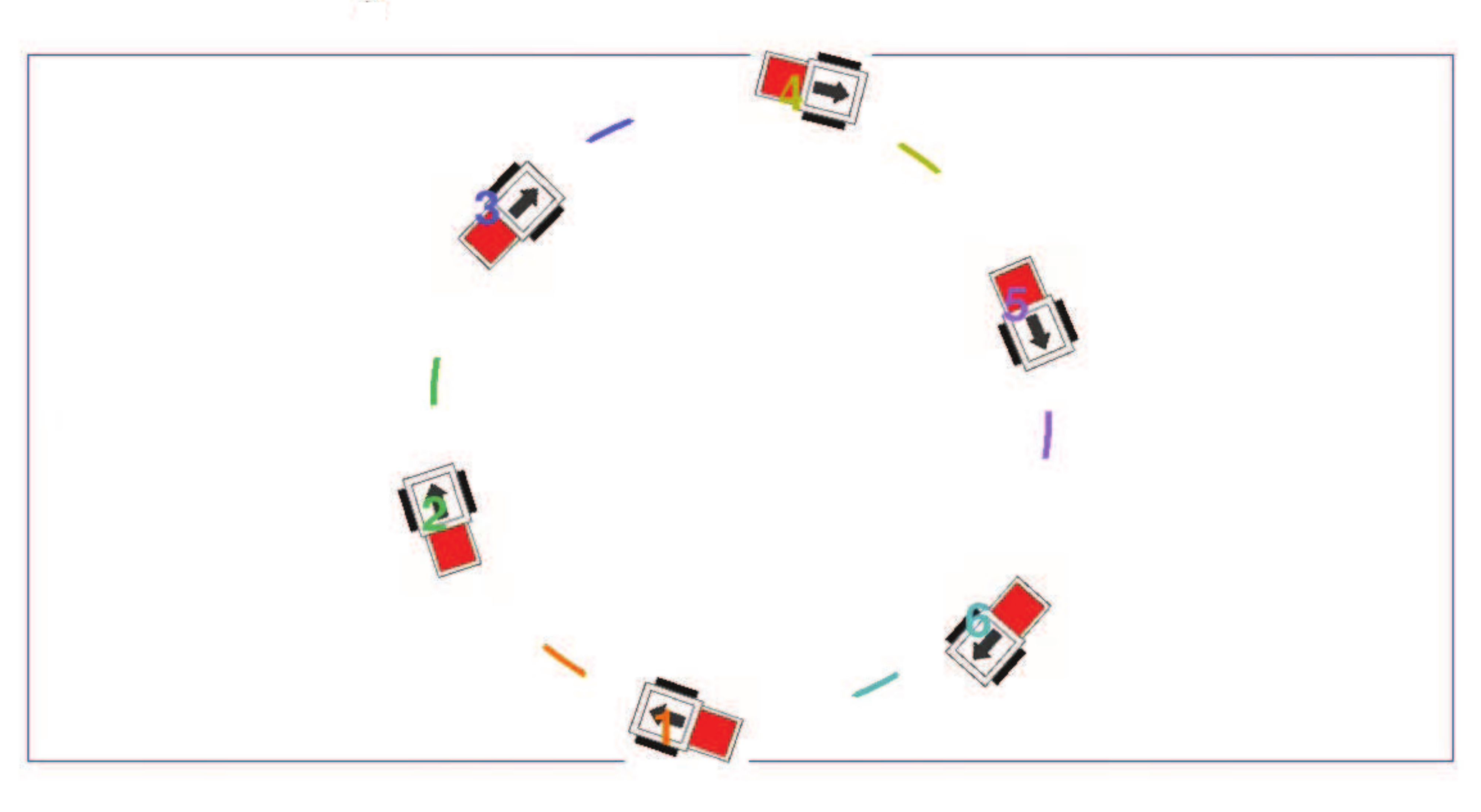}} && (b)\\
\raisebox{-.5\height}
{\includegraphics[width=0.40\textwidth]{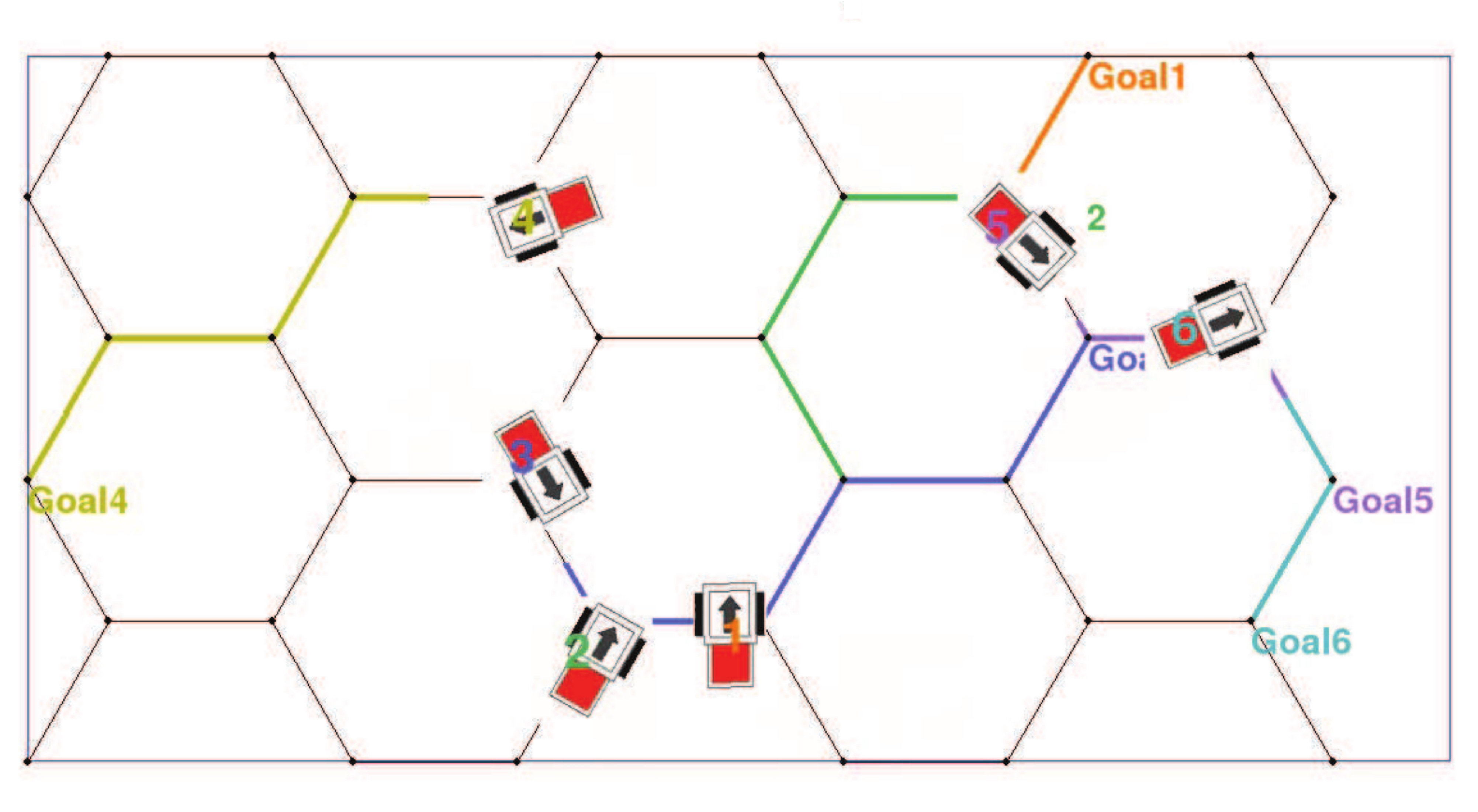}} && (c) \\ 
\raisebox{-.5\height}
{\includegraphics[width=0.40\textwidth]{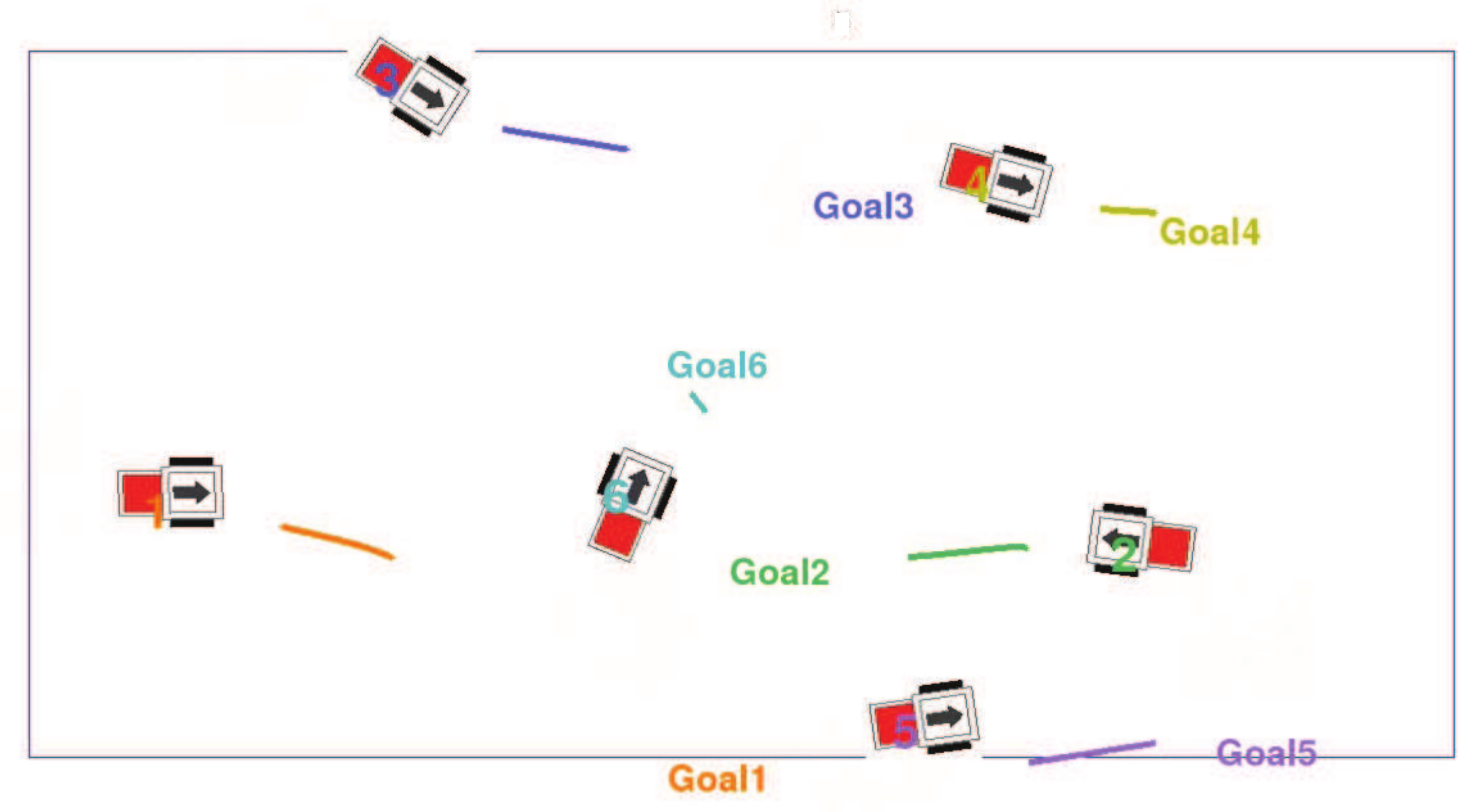}} && (d)\\
\end{tabular}
\end{center}
\caption{Snapshots of the GUI for different scenarios. (a) Vehicles 
following arbitrary paths hand-drawn by the end user. (b) Vehicles 
in formation, synchronously moving along the same circle. (c) Experiment
demonstrating the execution of a distance-optimal multi-robot path 
planning algorithm. (d) Vehicles coordinating using speed profiles generated 
by reciprocal velocity obstacle (RVO).}
\label{fig:scenarios}
\end{figure}

We highlight four application scenarios: {\em (i)} arbitrary path following
for an ``active'' vehicle, {\em (ii)} synchronized path following, 
{\em (iii)} running distance optimal multi-robot path
planning algorithms over a hexagonal grid, and {\em (iv)} distributed 
path planning using reciprocal velocity obstacles. GUI snapshots of these 
scenarios are provided in Fig.~\ref{fig:scenarios}. All these scenarios are 
included in the accompanying video, also available at 
\url{https://youtu.be/gXayWyRWDsw}. We intentionally shot the video as a 
real-time {\em long take} with no editing to demonstrate the reliability 
and robustness of \mvp when transitioning between different scenarios.

\subsection{Path Following}
\begin{figure}[ht!]
\begin{center}
\begin{overpic}[width=0.45\textwidth,tics=20]
{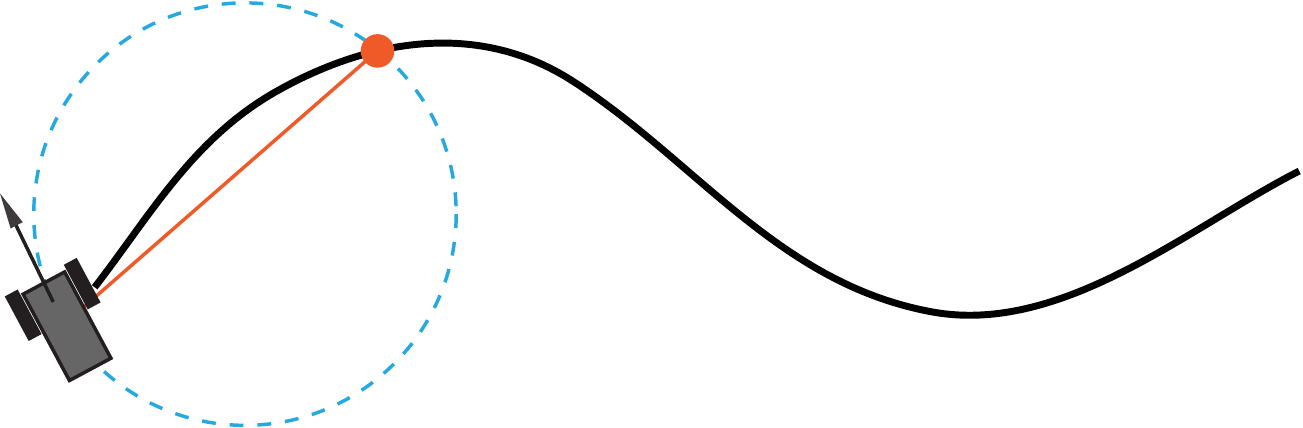}
\put(18,14){{\small $\ell$}}
\end{overpic}
\end{center}
\caption{Illustration of the pure pursuit algorithm for path following. 
Given an arbitrary curve defined by waypoints, the pure pursuit 
algorithm locates the first way point on the curve that is of some 
fixed distance $\ell$ from the vehicle location. Then, a circle is computed
that goes through the vehicle location and the fixed waypoint. The circle 
is also required to be tangential to the vehicle's current heading. The 
vehicle then attempts to follow the curvature determined by this circle 
by adjusting thrusts sent to its wheels.}
\label{fig:purepursuit}
\end{figure}

\mvp has built-in support for waypoint-based path tracking and following. 
At the single vehicle level, we adopt the pure pursuit 
\cite{wallace1985first,coulter1992implementation} algorithm 
that computes the desired curvature for a given look-ahead distance 
(see Fig.~\ref{fig:purepursuit}). The curvature then translates to desired
motor thrusts, allowing the vehicle to follow the input path 
(Fig.~\ref{fig:scenarios}(a)). 

In a multi-vehicle setup, several issues must be considered in \mvp: 
{\em (i)} the low-priced motors are inherently inaccurate in mapping input 
motor thrusts to output rotational speeds, {\em (ii)} uneven drag on the 
ground, {\em (iii)} camera lens induced workspace distortion, {\em (iv)} 
path curvature induced linear speed mismatch, and {\em (v)} motor speed 
variation due to battery drain variations. In computing motor thrusts for 
achieving desired curvature, \mvp limits the maximum thrust for each wheel 
motor at roughly $70\%$ of the maximum possible. This design choice leaves 
flexibility when it comes to synchronizing the movements of multiple 
vehicles. Roughly speaking, using waypoints with attached timestamps, \mvp 
dynamically adjust the maximum allowed motor thrusts, speeding up vehicles 
that fall behind and slowing down the ones that are ahead. With this 
mechanism, \mvp allows vehicles to accurately perform synchronized ``dance 
moves'' (Fig.~\ref{fig:scenarios}(b)). 

\subsection{Optimal Multi-Robot Path Planning}
To further verify the synchronous path following capability of \mvp, we 
integrated the optimal graph-based multi-robot path planning algorithm 
\cite{YuLav15TRO-II} into \mvp. Specifically, we tested the algorithm 
{\sc MinMaxDist} that minimizes the maximum distance traveled by any 
vehicle over a hexagonal grid. The size of the grid is determined 
using the vehicle footprint (see \cite{YuRus14ISRR}). For randomly 
generated goals for the fully distinguishable vehicles, we employ 
{\sc MinMaxDist} to compute the paths, turn them into waypoint-based,
time-synchronized trajectories, and then invoke the pure pursuit based 
path following API of \mvp to track these trajectories 
(Fig.~\ref{fig:scenarios}(c)). 

\subsection{Reciprocal Velocity Obstacles}
\mvp also supports the distributed  reciprocal velocity obstacle (RVO) 
based path planning algorithms. With minimal effort, we were able to 
integrate \mvp with the RVO2 library 
\cite{BergLinManocha08RVO,Snape2014DifferntialDrive}. Because RVO produces 
velocity profiles, we derive current vehicle velocities from vehicle 
locations. Then, the desired output velocities are simulated for a few 
steps to generate a set of paths with time synchronized waypoints. We 
may then invoke path following capability of \mvp to track these paths
repeatedly (Fig.~\ref{fig:scenarios}(d)). 

\section{Conclusion and Future Work}\label{section:conclusion}
In this paper, we summarized the design, implementation, and capabilities 
of \mvp. \mvp is intended to serve as a testbed for multi-robot 
planning and coordination algorithms and as an educational tool in the 
teaching of robotics subjects including mobile robots and multi-robot 
systems. Enabled by 3D-printing and the maker culture, \mvp is highly 
portable, readily affordable, low maintenance, and yet highly capable as
an open source multi-vehicle platform. 

\mvp will be continuously improved in its current and future iterations. 
On the hardware side, with the rapid development of IC technologies and 
improved design, the vehicles will become smaller, more accurate, and at 
the same time more affordable. In particular, we expect the release of a 
smaller vehicle costing around 35 USD in in the near future. We will also 
add on-board sensing capabilities to the vehicles while maintaining the
platform's affordability. On the software 
side, to improve the accuracy of vehicle state estimation, we are working
on a Extended Kalman Filter (EKF) to improve the sensing accuracy. 
Additional vehicle control APIs, including high level path planning with 
obstacle avoidance, will also be added to \mvp. Last, this open source 
effort hopes to solicit designs from all interested parties to make \mvp a 
community-based effort to promote robotics research and education. 

\bibliographystyle{IEEETran}
\bibliography{main,pubs}

\begin{thebibliography}{10}
\providecommand{\url}[1]{#1}
\csname url@samestyle\endcsname
\providecommand{\newblock}{\relax}
\providecommand{\bibinfo}[2]{#2}
\providecommand{\BIBentrySTDinterwordspacing}{\spaceskip=0pt\relax}
\providecommand{\BIBentryALTinterwordstretchfactor}{4}
\providecommand{\BIBentryALTinterwordspacing}{\spaceskip=\fontdimen2\font plus
\BIBentryALTinterwordstretchfactor\fontdimen3\font minus
  \fontdimen4\font\relax}
\providecommand{\BIBforeignlanguage}[2]{{%
\expandafter\ifx\csname l@#1\endcsname\relax
\typeout{** WARNING: IEEEtran.bst: No hyphenation pattern has been}%
\typeout{** loaded for the language `#1'. Using the pattern for}%
\typeout{** the default language instead.}%
\else
\language=\csname l@#1\endcsname
\fi
#2}}
\providecommand{\BIBdecl}{\relax}
\BIBdecl

\bibitem{YuLav15TRO-II}
J.~Yu and S.~M. LaValle, ``Optimal multi-robot path planning on graphs:
  Complete algorithms and effective heuristics,'' \emph{IEEE Transactions on
  Robotics}, to appear.

\bibitem{BergLinManocha08RVO}
J.~van~den Berg, M.~C. Lin, and D.~Manocha, ``Reciprocal velocity obstacles for
  real-time multi-agent navigation,'' in \emph{Proceedings IEEE International
  Conference on Robotics \& Automation}, 2008, pp. 1928--1935.

\bibitem{Snape2014DifferntialDrive}
J.~Snape, S.~J. Guy, J.~van~den Berg, and D.~Manocha, ``Smooth coordination and
  navigation for multiple differential-drive robots,'' in \emph{Experimental
  Robotics}.\hskip 1em plus 0.5em minus 0.4em\relax Springer, 2014.

\bibitem{mondada2009puck}
F.~Mondada, M.~Bonani, X.~Raemy, J.~Pugh, C.~Cianci, A.~Klaptocz, S.~Magnenat,
  J.-C. Zufferey, D.~Floreano, and A.~Martinoli, ``The e-puck, a robot designed
  for education in engineering,'' in \emph{Proceedings of the 9th conference on
  autonomous robot systems and competitions}, vol.~1, no.
  LIS-CONF-2009-004.\hskip 1em plus 0.5em minus 0.4em\relax IPCB: Instituto
  Polit{\'e}cnico de Castelo Branco, 2009, pp. 59--65.

\bibitem{rubenstein2014kilobot}
M.~Rubenstein, C.~Ahler, N.~Hoff, A.~Cabrera, and R.~Nagpal, ``Kilobot: A low
  cost robot with scalable operations designed for collective behaviors,''
  \emph{Robotics and Autonomous Systems}, vol.~62, no.~7, pp. 966--975, 2014.

\bibitem{pickem2016safe}
D.~Pickem, L.~Wang, P.~Glotfelter, Y.~Diaz-Mercado, M.~Mote, A.~Ames, E.~Feron,
  and M.~Egerstedt, ``Safe, remote-access swarm robotics research on the
  robotarium,'' \emph{arXiv preprint arXiv:1604.00640}, 2016.

\bibitem{Rey87}
C.~W. Reynolds, ``Flocks, herds, and schools: A distributed behavioral model,''
  \emph{Computer Graphics (ACM SIGGRAPH 87 Conf. Proc.)}, vol.~21, pp. 25--34,
  1987.

\bibitem{itseez2015opencv}
Itseez, ``Open source computer vision library,''
  \url{https://github.com/itseez/opencv}, 2015.

\bibitem{Lav06}
S.~M. LaValle, \emph{Planning Algorithms}.\hskip 1em plus 0.5em minus
  0.4em\relax Cambridge, U.K.: Cambridge University Press, 2006, also available
  at http://planning.cs.uiuc.edu/.

\bibitem{wallace1985first}
R.~Wallace, A.~Stentz, C.~E. Thorpe, H.~Maravec, W.~Whittaker, and T.~Kanade,
  ``First results in robot road-following.'' in \emph{IJCAI}, 1985, pp.
  1089--1095.

\bibitem{coulter1992implementation}
R.~C. Coulter, ``Implementation of the pure pursuit path tracking algorithm,''
  DTIC Document, Tech. Rep., 1992.

\bibitem{YuRus14ISRR}
J.~Yu and D.~Rus, ``An effective algorithmic framework for near optimal
  multi-robot path planning,'' in \emph{Proceedings International Symposium on
  Robotics Research}, 2015.

\end{thebibliography}

\end{document}